\documentclass[10pt,twocolumn,letterpaper]{article}

\usepackage[pagenumbers]{wacv} 

\usepackage{graphicx}
\usepackage{amsmath}
\usepackage{amssymb}
\usepackage{booktabs}
\usepackage[accsupp]{axessibility}  

\usepackage{algorithm2e}
\usepackage{algpseudocode}
\RestyleAlgo{ruled}
\LinesNumbered
\SetKw{Continue}{continue}

\usepackage{floatrow}
\newfloatcommand{capbtabbox}{table}[][\FBwidth]

\usepackage{xcolor}
\usepackage{multirow}
\usepackage{adjustbox}
\usepackage{subcaption}
\usepackage{transparent}
\usepackage[accsupp]{axessibility} 
    
\definecolor{todocolour}{RGB}{0,165,135}

\usepackage[pagebackref,breaklinks,colorlinks]{hyperref}

\usepackage[capitalize]{cleveref}
\crefname{section}{Sec.}{Secs.}
\Crefname{section}{Section}{Sections}
\Crefname{table}{Table}{Tables}
\crefname{table}{Tab.}{Tabs.}

\definecolor{rebuttalcolor}{RGB}{0,165,135}

\def\wacvPaperID{53} 
\def\confName{WACV}
\def\confYear{2024}

\begin{document}

\title{Continual atlas-based segmentation of prostate MRI}


\author{Amin Ranem$^{1}$\\
\and 
Camila Gonz\'alez$^{2}$\\
\and
Daniel Pinto dos Santos$^{3}$\\
\and
Andreas M. Bucher$^{4}$\\
\and
Ahmed E. Othman$^{5}$\\
\and
Anirban Mukhopadhyay$^{1}$\\
\and
{\small$^{1}$ Technical University of Darmstadt}, 
{\small$^{2}$ Stanford University}, 
{\small$^{3}$ University of Cologne},\\
{\small$^{4}$ University of Frankfurt},
{\small$^{5}$ University Medical Center Mainz}\\
{\tt\small amin.ranem@gris.tu-darmstadt.de}}

\maketitle

\begin{abstract}
Continual learning (CL) methods designed for natural image classification often fail to reach basic quality standards for medical image segmentation. Atlas-based segmentation, a well-established approach in medical imaging, incorporates domain knowledge on the region of interest, leading to semantically coherent predictions. This is especially promising for CL, as it allows us to leverage structural information and strike an optimal balance between model rigidity and plasticity over time. When combined with privacy-preserving prototypes, this process offers the advantages of rehearsal-based CL without compromising patient privacy. We propose \textbf{Atlas Replay}, an atlas-based segmentation approach that uses prototypes to generate high-quality segmentation masks through image registration that maintain consistency even as the training distribution changes. We explore how our proposed method performs compared to state-of-the-art CL methods in terms of knowledge transferability across seven publicly available prostate segmentation datasets. Prostate segmentation plays a vital role in diagnosing prostate cancer, however, it poses challenges due to substantial anatomical variations, benign structural differences in older age groups, and fluctuating acquisition parameters. Our results show that Atlas Replay is both robust and generalizes well to yet-unseen domains while being able to maintain knowledge, unlike end-to-end segmentation methods. Our code base is available under \url{https://github.com/MECLabTUDA/Atlas-Replay}.
\end{abstract}

\section{Introduction}
Continual learning (CL) plays a crucial role in safety-critical applications of Deep Learning (DL), particularly in healthcare. In such domains, models must continually adapt to data drift over time while maintaining high performance on older data, even in cases where direct access to part of the data is restricted for privacy reasons. The objective of CL is to train a model that demonstrates high performance on sequentially arriving datasets, despite the constrained timeframe during which the datasets are accessible. Achieving this objective is challenging, as approaches tend to fall into one of the extremes: either suffering from \emph{catastrophic forgetting} \cite{kirkpatrick2017overcoming} by being too plastic, or \emph{unable to learn} new tasks by being too rigid. Additionally, some methods exhibit linear growth in training time and resource requirements as the number of training tasks increases. CL models are further susceptible to domain shifts over time \cite{gonzalez2022task}, increasing the amount of catastrophic forgetting even further. While various strategies have been proposed for CL, they often fail to perform well with medical data \cite{derakhshani2022lifelonger}, resulting in segmentations that do not meet basic semantic standards. Therefore, striking a delicate balance between preserving previous knowledge and maintaining the necessary plasticity to learn new tasks is a key requirement when training in a continual fashion \cite{de2021continual, hadsell2020embracing}.

As we have domain knowledge on the geometry of the organ to be segmented, we go back to the roots of medical image segmentation and contextualize atlas-based segmentation \cite{lorenzo2002atlas}, as an alternative to end-to-end CL segmentation pipelines. In the context of CL, the access to structural information is key for achieving accurate and semantically coherent predictions. In atlas-based segmentation, registration is used to adapt the best-fitting labeled mask from an atlas (i.e. a pre-defined set of reference images) to a patient scan.  The natural reliance of atlas-based approaches on geometrical aspects makes it ideally applicable for CL setups as the structural information can be leveraged to extract domain knowledge.

\begin{figure*}[htb]
    \begin{subfigure}[t]{.235\linewidth}
        \centering\includegraphics[trim=0cm 10cm 26cm 2.3cm, clip, width=\textwidth]{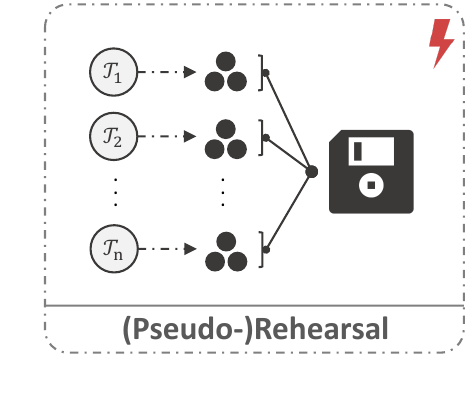}
        \caption{Privacy concerns}
        \label{fig:priv_conc}
    \end{subfigure}
    \begin{subfigure}[t]{.470\linewidth}
        \centering\includegraphics[trim=8cm 10cm 10.25cm 2.3cm, clip, width=\textwidth]{Images/fig_intro.pdf}
        \caption{Rigidity/plasticity trade-off -- No semantic coherence}
        \label{fig:trade}
    \end{subfigure}
    \begin{subfigure}[t]{.235\linewidth}
        \centering\includegraphics[trim=23.5cm 10cm 2.5cm 2.3cm, clip, width=\textwidth]{Images/fig_intro.pdf}
        \caption{Dependent on capacity}
        \label{fig:grow}
    \end{subfigure}
    \caption{Demonstration of different problems CL methods face.}
    \label{fig:intro}
\end{figure*}

CL for end-to-end models can be generally classified into one of the following three categories: (1) (pseudo-) rehearsal, (2) regularization/distillation, and (3) expansion. Figure \ref{fig:intro} demonstrates the practical downsides of each strategy.

The most successful CL methods follow some type of \textit{replay} or \textit{rehearsal}, which involves storing samples to interleave them during later training \cite{rebuffi2017icarl,wu2019large}. Long-term storage of patient scans, however, violates data protection regulations \cite{european_commission_regulation_2016}. Distillation methods such as \emph{PLOP} \cite{douillard2021plop} do not store data directly but rather distill knowledge from previously trained models. Yet recent work \cite{gonzalez2022lifelong} shows that the additional computational burden associated with the pseudo-data generation hinders their use with high-dimensional medical images. Regularization methods \cite{chaudhry2018riemannian,kirkpatrick2017overcoming} on the other hand work by penalizing severe shifts from the previously-learned parameter space. These approaches have lower resource requirements, but disappointing performance across all tasks and merely allow a trade-off between rigidity and plasticity. Models with high plasticity increase the ability to learn new information, whereas rigid models maintain more knowledge from previous tasks and prevent catastrophic forgetting. Additionally, models trained end-to-end in a sequential manner with regularization or distillation approaches tend to generate predictions with no semantic coherence. 
Expansion techniques maintain stable performance across all tasks but grow the model size with the number of tasks \cite{gonzalez2022task,hung2019compacting}. Other works \cite{memmel2021adversarial,ozgun2020importance,ranem2022continual} propose architectural modifications that are useful in certain settings but imply a high computational overhead. Instead of relying on end-to-end CL, we propose a modular atlas-based method for continual segmentation. \emph{Atlas Replay} leverages structural information to maintain knowledge over time with the benefits of rehearsal while preserving patient privacy and avoiding model growing.

\textbf{Atlas Replay} generates \emph{prototypes} built from a set of patient scans and the \emph{VoxelMorph} registration framework \cite{balakrishnan2019voxelmorph} to perform registration between a specific patient scan and a prototype. A prototype is a combination of multiple patient scans -- and corresponding segmentation masks -- that disallows the direct identification of a subject. Prototypes are registered to new images to generate segmentation masks for new patients during deployment.

We introduce an approach to build an atlas of prototypes, \textit{irrespective of the anatomy}; and \textit{propose a CL method} to perform atlas-based segmentation which maintains model plasticity while preserving previous knowledge and outperforms state-of-the-art (SOTA) end-to-end continual segmentation approaches. We can consider using a stored atlas to generate segmentation masks as a form of pseudo-rehearsal that maintains relevant information from previous examples without storing actual patient images.

The contributions of this work are three-fold. Our proposed approach:
\begin{itemize}
    \item succesfully builds privacy-preserved prototypes and therefore being more protective towards patient privacy compared to traditional replay-based methods,
    \item leverages prototypes and structural information to maintain knowledge over time,
    \item benefits from a rehearsal based approach and VoxelMorph to achieve stable performances for continual image segmentation.
\end{itemize}

To validate our method, we investigate the problem of prostate segmentation in T2-weighted Magnetic Resonance Images (MRI), which is an important step in the diagnosis and treatment of prostate cancer \cite{tempany2018role}. Prostates have relatively static shapes for which domain knowledge can be leveraged over time. Variations in imaging protocol, such as the diminishing use of endorectal coil over time \cite{lee2022prostate}, lead to domain shift. Such shifts clearly state the importance of CL from a clinical perspective.

With the introduction of Atlas Replay, we pave the path for integrating atlas-based methods into the realm of CL. We demonstrate that established conventional approaches like atlas-based segmentation, which have fallen into a certain neglect due to the current DL era, can be effectively utilized in dynamic clinical setups. This success can be attributed to the use of structural information, which showcases highly favorable outcomes in CL scenarios.

\section{Methodology}
Traditional end-to-end segmentation methods in CL typically struggle to strike a suitable balance between rigidity and plasticity, which presents a significant drawback in this field. Such models often face a trade-off where they either fail to acquire new knowledge by retaining information from previously seen data, or overly prioritize recent cases, leading to catastrophic forgetting and predictions lacking semantic coherence. Recognizing the importance of structural information for achieving a proper trade-off between rigidity and plasticity in CL setups, we turn to atlas-based segmentation as a solution. Most atlas-based segmentation methods are based on traditional (non-DL) techniques \cite{aljabar2009multi,lorenzo2002atlas}. We combine the advantages of DL with the ability to leverage domain knowledge of atlas-based segmentation for CL. An \textit{atlas} $\Lambda$ is a manually labeled set of patient scans \cite{rohlfing2005quo}. Atlas-based segmentation is the \textit{modular process} of registering an image from the atlas to a new scan to directly generate an accurate segmentation by transforming the respective mask in the same manner \cite{lorenzo2002atlas}, thus leveraging structural information. This approach has been successfully applied to prostate cancer \cite{aoyama2021comparison}, heart regions \cite{ghosh2021multi}, brain tissue \cite{aljabar2009multi} and aortic tissue \cite{isgum2009multi} for MRI scans.

\paragraph{Fundamentals}
We start by introducing some key terminology: $\Omega \subset \mathbb{R}^3$ defines a 3D spatial domain.
$\mathcal{T}_p \subset \Omega_{\mathcal{T}}$ is referred to as dataset $p$ and consists of $m \{(f^i_m, f^s_m)\}$ pairs, where $f^i$ is a patient scan and $f^s$ the corresponding segmentation mask. Stage $x$ in a continual setup defines the process of training the model on dataset $x$ after it has been trained on all previous $\{1, \dots, x-1\}$ tasks. 
$\Omega_{\mathcal{T}}$ is a set of datasets and $\Omega_{\mathcal{P}}$, a set of prototypes. A prototype $\mathcal{P}_k$ is a tuple $\mathcal{P}_k = (\mathcal{P}^i_k, \mathcal{P}^s_k)$, where $\mathcal{P}^i_k$ is the scan and $\mathcal{P}^s_k$ the corresponding segmentation mask. We define our set of prototypes $\Omega_{\mathcal{P}}$ as \textit{privacy-preserving representations}.

\emph{VoxelMorph} \cite{balakrishnan2019voxelmorph} is a popular framework for medical image registration. The underlying architecture is a simple U-Net, to which additional convolutional layers are added to generate a deformation field $\phi$. The network is trained by penalizing the difference between the warped moving image using $\phi$ and the target image. By using $\phi$ to alter the prototype segmentation mask, one can utilize the framework for atlas-based segmentation given a specific atlas $\Lambda$.

\paragraph{Prototype building}
{We create four distinct prototypes based on the coil type used during acquisition. For each prototype $\mathcal{P}_k$, we extract $r$ random samples from the associated dataset $\mathcal{T}_p$, a subset of $\Omega_{\mathcal{T}}$. In this study, we set $r=7$, corresponding to the size of the smallest training set. To evaluate the performance between prototypes, we allocate three datasets from our data corpus to validate the inter-prototype performance in Section \ref{ssec:ablation_prot}.

First, $r$ random images are selected from the training dataset $\mathcal{T}_{p}$. The first image $I$ represents the initial prototype $\mathcal{P}_k$. For each following image $I^*$, we refine $\mathcal{P}_k$ through rigid alignment using \emph{SimpleITK} \cite{lowekamp2013design, yaniv2018simpleitk}. We then compute the average of $\mathcal{P}_k$ and $I^*$ to update the prototype, including the corresponding segmentation masks. The final $\mathcal{P}k$ is the prototype for dataset $\mathcal{T}{p}$. Taking the average over multiple scans hinders the recovery of patient-sensitive information during storing or interleaving, as illustrated in Figure \ref{fig:prot_building}.

\begin{figure}[htbp!]
    \centering
    \includegraphics[trim=0 10cm 22cm 0, clip, width=0.95\textwidth]{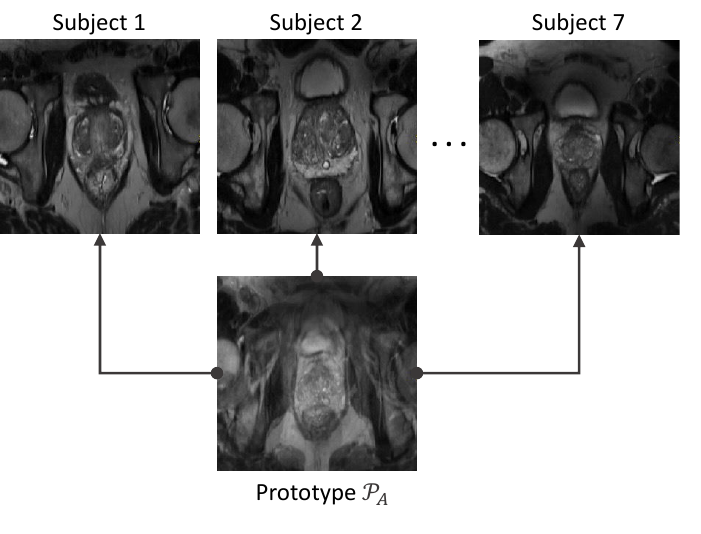}
    \caption{Illustration of prototype $\mathcal{P}_{A}$.}
    \label{fig:prot_building}
\end{figure}

Figure \ref{fig:prot_building} shows the difference between the final prototype $\mathcal{P}_{D}$ and three out of seven subject scans that were used to build the prototype. The intensity distribution for every dataset is shown in Figure \ref{fig:hists}. We asses the effectiveness of our prototypes in maintaining privacy by performing a user study among senior radiologists, with more than 10 years of experience, in Section \ref{ssec:user}. 

\begin{figure}[htb]
    \centering
    \includegraphics[trim=0 8cm 19.5cm 0, clip, width=\textwidth]{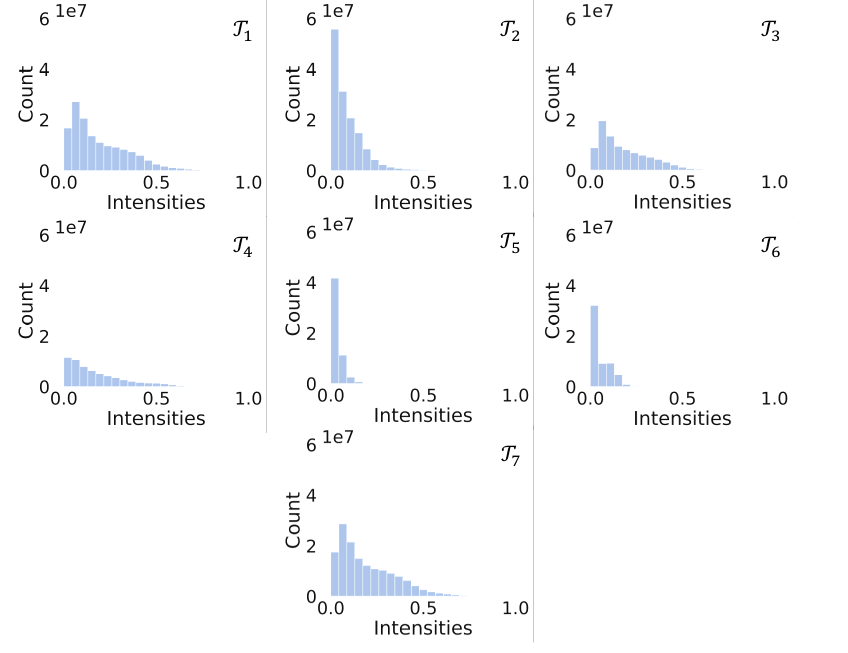}
    \caption{Intensity histograms for every dataset.}
    \label{fig:hists}
\end{figure}

\paragraph{Prototype registration}
A patient scan $f^{i}$ represents the fixed image that is registered to the best-fitting prototype $\mathcal{P}_{k}$ -- the moving image -- using \textit{VoxelMorph}. The deformation field $\phi$ from the network is then used to warp the prototypes` segmentation mask: $\mathcal{P}_{k}^{s} \circ \phi$. As illustrated in Figure \ref{fig:training_pip}, our method only stores \textit{prototypes} that preserve key information and therefore preserves patient privacy better than storing actual subjects.

\begin{figure*}[htb]
    \centering
    \includegraphics[trim=2cm 6.85cm 0 0, clip, width=\textwidth]{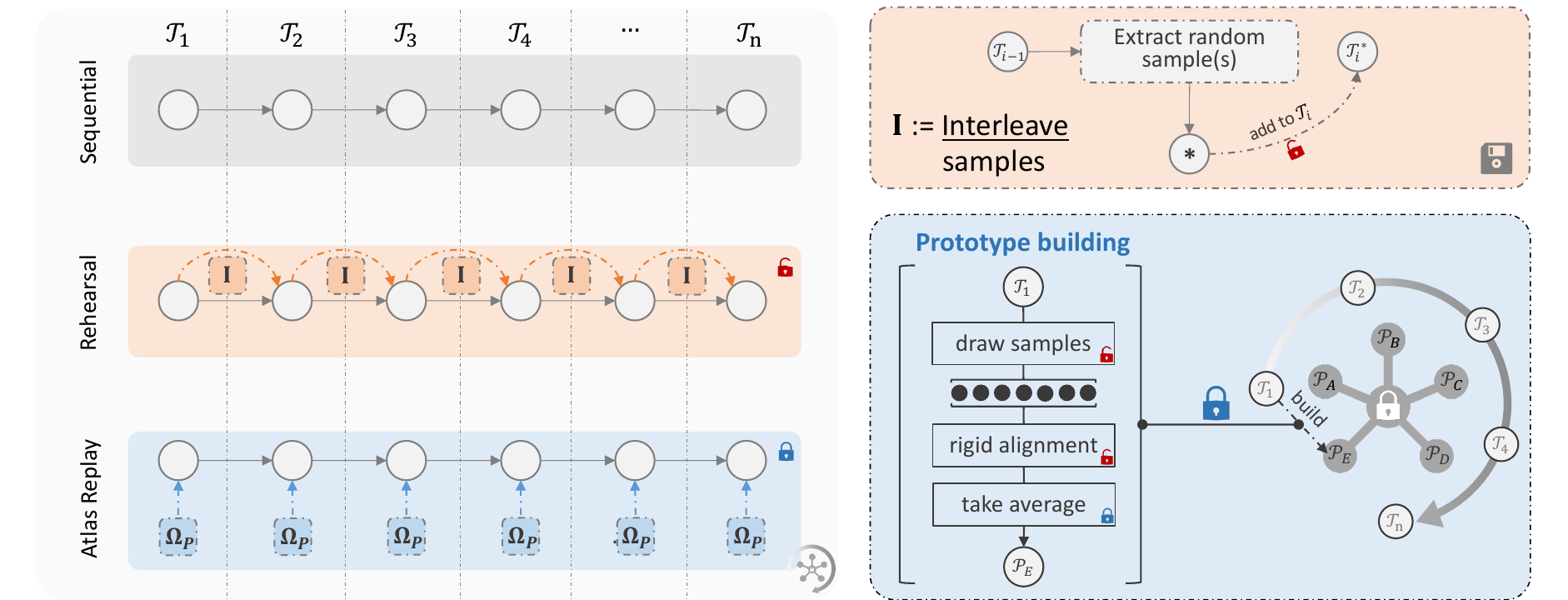}
    \caption{Comparison of sequential training on $n$ tasks against traditional rehearsal and \emph{Atlas Replay} in terms of privacy preservation; $\Omega_{\mathcal{P}}$ represents the stored prototypes.}
    \label{fig:training_pip}
\end{figure*}

\paragraph{Training continually using Atlas Replay}
Algorithm \ref{alg:train} demonstrates a continual training setup in simplified pseudo-code for our proposed Atlas Replay approach using the VoxelMorph (VxM) Framework.

\begin{algorithm}[htp]
\caption{Training using Atlas Replay}
\label{alg:train}
\KwIn{Datasets to train on $\{\mathcal{T}_{p}\}_{p \leq \lvert \Omega_{\mathcal{T}} \rvert}$}
\KwOut{Trained model weights $\theta$}
\tcp{Initialize $\mathcal{M}_\theta$}
$\theta \gets initializeModel()$ 

\tcp{Build $k=4$ prototypes}
$\{\mathcal{P}_k\} \gets buildPrototypes \left(\{\mathcal{T}_{p}\}_{p \leq \lvert \Omega_{\mathcal{T}} \rvert}, k=4 \right)$

\tcp{Select prototype for $\mathcal{T}_1$}
$\mathcal{P} \in \{\mathcal{P}_k\}$

\tcp{Train with VxM}
$\theta \gets \text{VxM}(\theta, \mathcal{T}_1, \mathcal{P})$

\For{$i\leftarrow 2$ \KwTo $\lvert \Omega_{\mathcal{T}} \rvert$}{

    \tcp{Select prototype for $\mathcal{T}_i$}
    $\mathcal{P} \in \{\mathcal{P}_k\}$
    
    \tcp{Train with VxM}
    $\theta \gets \text{VxM}(\theta, \mathcal{T}_i, \mathcal{P})$
}
\end{algorithm}

\section{Experimental Setup}
In this section, we briefly describe our corpus of publicly available datasets and report essential aspects of our experimental setup.

\paragraph{Data corpus}
Our prostate data corpus $\Omega_{\mathcal{T}}$ consists of seven publicly available datasets \cite{litjens2014evaluation, NCI-ISBI, lemaitre2015computer, liu2020saml, liu2020ms, antonelli2021medical}. 
Table \ref{tab:data} shows the number of cases in each dataset and the random 80:20 train/validation split. The splits along with our code base is accessible under \url{https://github.com/MECLabTUDA/Atlas-Replay}.

\begin{table}[htp]
\centering
\begin{adjustbox}{max width=\linewidth}
{\begin{tabular}{lcccc}
\toprule
 Dataset & Task ID & \# Cases (train, val) & Vendor & Source \\
\midrule \midrule
RUNMC & $\mathcal{T}_{1}$ & 30 -- (24, 6) & Siemens & \multirow{2}{*}{\cite{NCI-ISBI}}\\
BMC & $\mathcal{T}_{2}$ & 30 -- (24, 6) & Philips & \\
HCRUDB & $\mathcal{T}_{3}$ & 19 -- (15, 4) & Siemens & \cite{lemaitre2015computer} \\
UCL & $\mathcal{T}_{4}$ & 13 -- (10, 3) & Siemens & \multirow{3}{*}{\cite{litjens2014evaluation}}\\
BIDMC & $\mathcal{T}_{5}$ & 12 -- (9, 3) & GE & \\
HK & $\mathcal{T}_{6}$ & 12 -- (9, 3) & Siemens & \\ \hline
DecathProst & $\mathcal{T}_{7}$ & 32 -- (25, 7) & Unknown & \cite{antonelli2021medical}\\
\bottomrule\\
\end{tabular}}
\end{adjustbox}
\caption{Characteristics of our prostate data corpus; including the vendor of the acquisition device.}
\label{tab:data}
\end{table}

\paragraph{Prototypes}
To give the reader a proper understanding of our set of prototypes $\Omega_{\mathcal{P}}$, these are illustrated in Figure \ref{fig:prots}.

\begin{figure}[htbp!]
    \begin{subfigure}[t]{.235\linewidth}
        \centering\includegraphics[width=\textwidth]{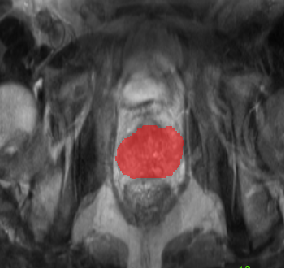}
        \caption{$\mathcal{P}_{A}$}
        \label{fig:no}
    \end{subfigure}
    \begin{subfigure}[t]{.235\linewidth}
        \centering\includegraphics[width=\textwidth]{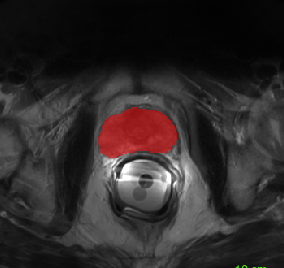}
        \caption{$\mathcal{P}_{B}$}
        \label{fig:light}
    \end{subfigure}
    \begin{subfigure}[t]{.235\linewidth}
        \centering\includegraphics[width=\textwidth]{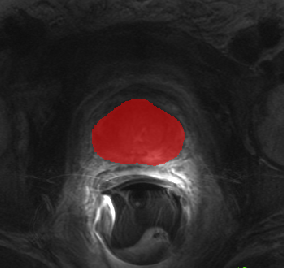}
        \caption{$\mathcal{P}_{C}$}
        \label{fig:mid}
    \end{subfigure}
    \begin{subfigure}[t]{.235\linewidth}
        \centering\includegraphics[width=\textwidth]{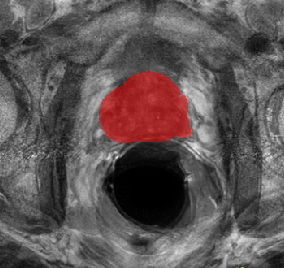}
        \caption{$\mathcal{P}_{D}$}
        \label{fig:dark}
    \end{subfigure}
    \caption{Prototypes and their overlapping segmentation mask. For every prototype, a random slice was selected.}
    \label{fig:prots}
\end{figure}

\paragraph{Training setup}
The VoxelMorph framework \cite{balakrishnan2019voxelmorph} with default optimizer and learning rate is used for all registration experiments. To ensure a fair comparison between the network's performance, the same underlying U-Net architecture is used for end-to-end segmentation. Registration models are trained for 250 epochs using a composition of the Normalized Cross Correlation ($\mathcal{L}_{NCC}$), Cross-Entropy (CE, $\mathcal{L}_{CE}$), and a gradient smoothing loss for the deformation field ($\mathcal{L}_{smooth}$). The total registration loss is calculated in the following way:
\begin{align}
\label{align:loss}
    \mathcal{L}_{reg} = \mathcal{L}_{NCC} + 2 \cdot \mathcal{L}_{CE} + \mathcal{L}_{smooth}
\end{align}
Based on our ablation results from Section \ref{ssec:loss}, we use the double-weighted CE loss setup for all our VoxelMorph experiments. Segmentation networks are trained for 250 epochs with $\mathcal{L}_{CE}$ only. All experiments were carried out on a GeForce RTX 3090 GPU (24 GB).

\paragraph{Metrics}
We report the mean Dice and standard deviation across the test images from all tasks as well as average backward (BWT) and forward (FWT) transferability \cite{lopez2017gradient, diaz2018don}. BWT indicates the amount of maintained knowledge on test samples $(f^i_m, f^s_m) \subset \mathcal{T}_j$ during training on different stages $\{\mathcal{T}_p\}_{p \leq \lvert \Omega_{\mathcal{T}} \rvert}\; ; \;j < p$ over time. FWT on the other hand measures the impact of the current training stage $\{\mathcal{T}_p\}_{p \leq \lvert \Omega_{\mathcal{T}} \rvert}$ on test data $(f^i_m, f^s_m) \subset \mathcal{T}_j\; ; \;j > p$ from an untrained stage.

Let $\mathcal{T}_{p}$ be a specific task:
\begin{align}
\label{eqn:B}
    \text{BWT}\left( \mathcal{T}_{p}\right) &=
    \text{Dice}\left(\mathcal{M}_{\left[ \mathcal{T}_{1}, \dots, \mathcal{T}_{p}, \dots, \mathcal{T}_{n}\right]}, \mathcal{T}_{p}\right) \nonumber\\
    &\:- \text{Dice}\left(\mathcal{M}_{\left[ \mathcal{T}_{1}, \dots, \mathcal{T}_{p}\right]}, \mathcal{T}_{p}\right),
\end{align}
where $\mathcal{M}_{\left[ \mathcal{T}_{1}, \dots, \mathcal{T}_{p}\right]}$ is a network trained on stages $\{1, \dots, p\} \leq \lvert \Omega_{\mathcal{T}} \rvert$ and $\text{Dice}(\mathcal{M}_{\left[ \mathcal{T}_{1}, \dots, \mathcal{T}_{j}\right]}, \mathcal{T}_{p})$ indicates the S\o{}rensen–Dice coefficient from a network trained on stages $\{1, \dots, j\}$ evaluated on dataset $p$. FWT is defined as: 
\begin{align}
\label{eqn:F}
    \text{FWT}\left( \mathcal{T}_{p}\right) &= \text{Dice}\left(\mathcal{M}_{\left[ \mathcal{T}_{1}, \dots, \mathcal{T}_{p-1}\right]}, \mathcal{T}_{p}\right) \nonumber\\
    &\:- \text{Dice}\left(\mathcal{M}_{\left[\mathcal{T}_{p}\right]}, \mathcal{T}_{p}\right).
\end{align}
FWT for the last model state as well as BWT for the first model state are not defined. Models with high plasticity are able to learn new knowledge and achieve higher FWT, whereas models that maintain most knowledge from previous tasks obtain a higher BWT.

To validate our user study, we report sensitivity, specificity, precision (positive predicted value), and the Matthews Correlation Coefficient (MCC) \cite{trevethan2017sensitivity, chicco2020advantages, matthews1975comparison}. We calculate the MCC to provide an overall assessment of the classification performance with respect to both, true negative and true positive rates.

\paragraph{Baselines}
We compare Atlas Replay to end-to-end sequential training, the upper bound of simple replay training, which requires the storage of actual patient scans, and four popular CL methods: Elastic Weight Consolidation (EWC) \cite{kirkpatrick2017overcoming}, Riemannian Walk (RWalk) \cite{chaudhry2018riemannian}, Incremental Learning Techniques (ILT) \cite{michieli2019incremental} using distillation on the output (KD), intermediate (MSE), or both (KD, MSE) layers and Bias Correction (BiC) \cite{wu2019large}. A hyperparameter search is conducted for EWC and RWalk, and the best settings are used of $\lambda = 2.2$ (EWC), $\alpha = 0.9$, and $\lambda = 1.7$ (RWalk). For ILT, we used the default distillation parameter $\lambda_D=1$. Since BiC is a rehearsal-based method, from each task, seven samples were interleaved.

\section{Results}
We present a comprehensive evaluation of various aspects of our work. We start by analyzing the performance of Atlas Replay compared to sequential training, EWC, RWalk, BiC, and rehearsal training. We then proceed with a qualitative temporal evaluation in Section \ref{ssec:qual}. Section \ref{ssec:user} explores the effectiveness of privacy-preserving prototypes through a user study conducted with senior radiologists and computer scientists. We assess the inter-prototype performance during inference in \ref{ssec:ablation_prot}. Furthermore, we conduct a loss ablation study in \ref{ssec:loss} and compare the U-Net's end-to-end segmentation with our atlas-based segmentation approach in \ref{ssec:gener}. The results are based on the hyperparameter search we conducted in Section \ref{ssec:param}. These evaluations provide multiple insights into the performance, robustness, and effectiveness of the proposed approach and prototype generation strategy.

\subsection{Continual learning performance}
\label{ssec:cl_res}
In this section, we compare Atlas Replay to training a U-Net model sequentially alongside the continual learning methods EWC, RWalk, ILT and BiC. Further, we compare against the upper bound of rehearsal training (storing seven samples from each task).

\begin{figure}[htb]
    \begin{subfigure}[t]{.31\linewidth}
        \centering\includegraphics[trim=0 0 0 0.7cm, clip,width=\textwidth]{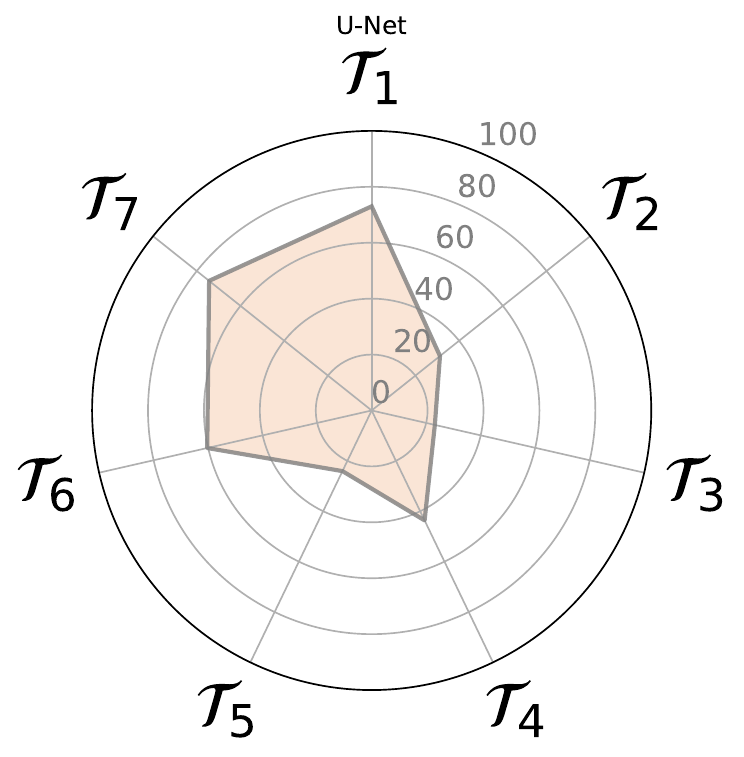}
        \caption{Sequential}
        \label{fig:spider_seq}
    \end{subfigure}
    \begin{subfigure}[t]{.31\linewidth}
        \centering\includegraphics[trim=0 0 0 0.7cm, clip,width=\textwidth]{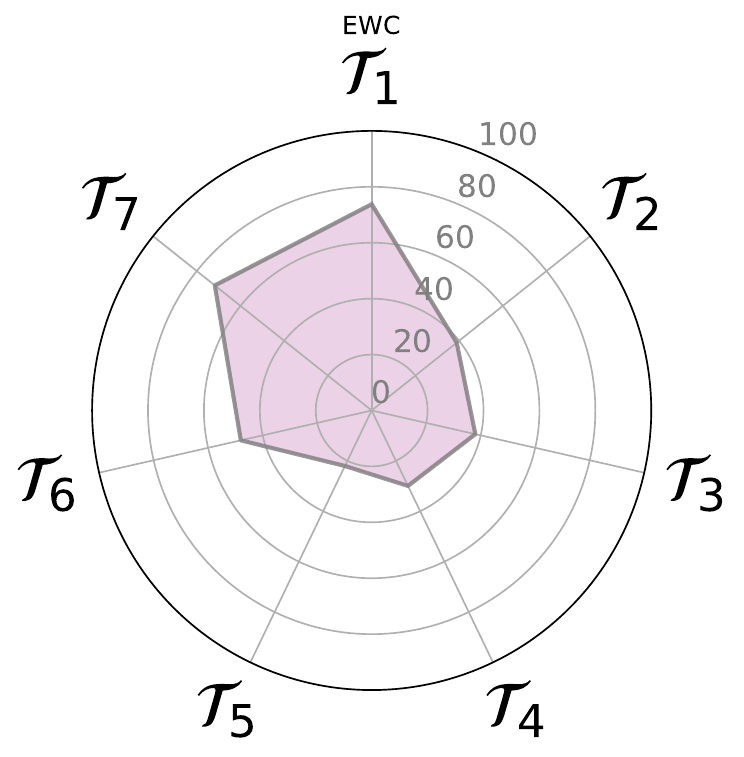}
        \caption{EWC}
        \label{fig:spider_ewc}
    \end{subfigure}
    \begin{subfigure}[t]{.31\linewidth}
        \centering\includegraphics[trim=0 0 0 0.7cm, clip,width=\textwidth]{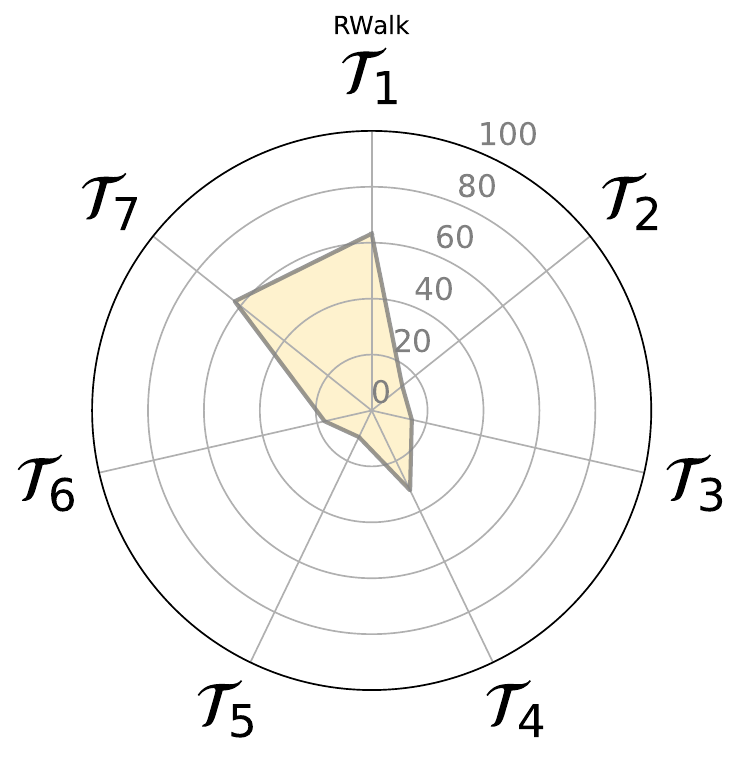}
        \caption{RWalk}
        \label{fig:spider_rwalk}
    \end{subfigure}
    \begin{subfigure}[t]{.31\linewidth}
        \centering\includegraphics[trim=0 0 0 0.7cm, clip,width=\textwidth]{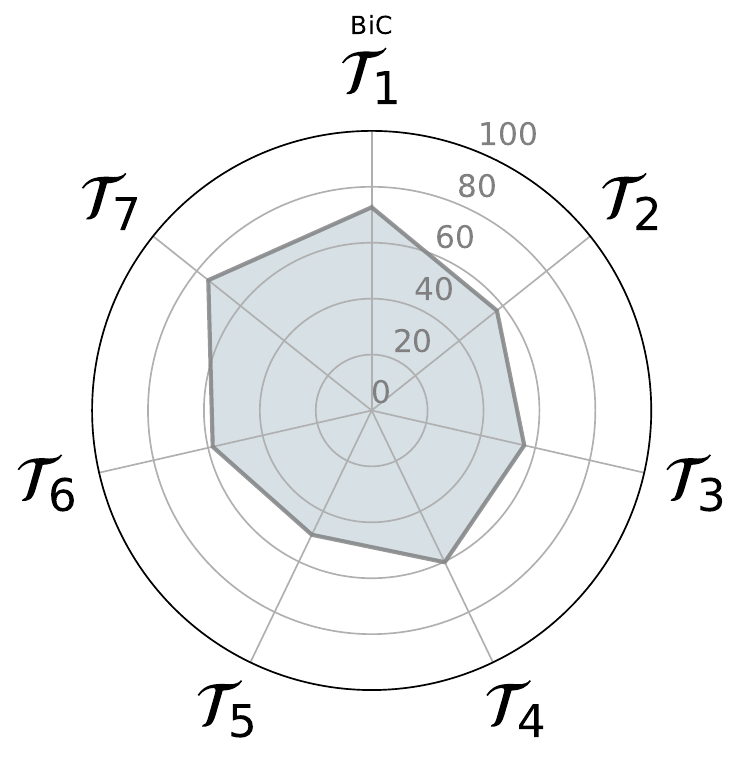}
        \caption{BiC}
        \label{fig:spider_bic}
    \end{subfigure}
    \begin{subfigure}[t]{.31\linewidth}
        \centering\includegraphics[trim=0 0 0 0.7cm, clip,width=\textwidth]{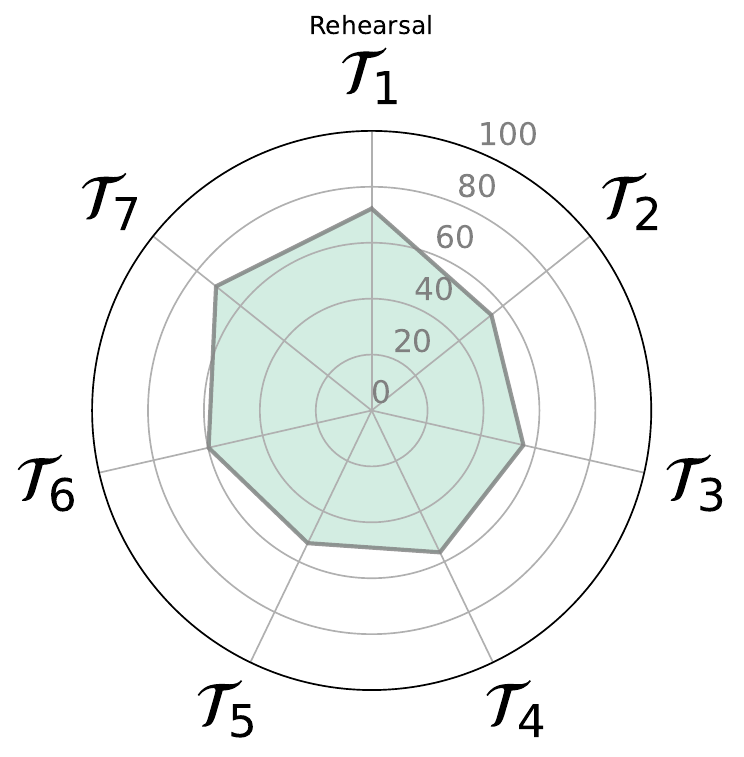}
        \caption{Rehearsal}
        \label{fig:spider_reh}
    \end{subfigure}
    \begin{subfigure}[t]{.31\linewidth}
        \centering\includegraphics[trim=0 0 0 0.7cm, clip,width=\textwidth]{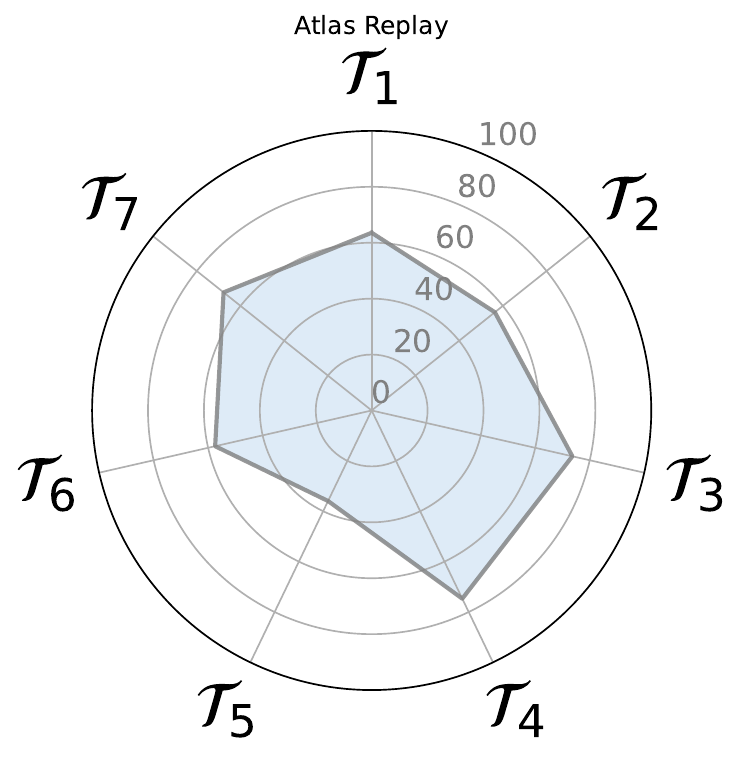}
        \caption{Atlas Replay}
        \label{fig:spider_ours}
    \end{subfigure}
    \caption{CL performance as Dice; the larger the area the better.}
    \label{fig:spiders}
\end{figure}

Figure \ref{fig:spiders} shows the mean Dice scores of the final networks evaluated across all seven datasets. EWC (\ref{fig:spider_ewc}) achieves more or less the same Dice performance across the tasks as the simple sequential setup (\ref{fig:spider_seq}) showing slight shifts in BWT and FWT as seen in Table \ref{tab:CL_perf}. RWalk, on the other hand, achieves a better BWT, meaning it maintains more knowledge. However, this comes at the cost of a lower mean Dice and FWT (\ref{fig:spider_rwalk}). This indicates that even after tuning the hyperparameters, end-to-end state-of-the-art CL methods under-perform for medical image segmentation. BiC on the other hand preserves previous knowledge the best (BWT), but at the cost of plasticity loss (FWT), Table \ref{tab:CL_perf}. Atlas Replay (\ref{fig:spider_ours}) is the only method that maintains good performance over all seen tasks. BWT is higher than for the rehearsal upper bound, as indicated in \ref{fig:spider_bic} to \ref{fig:spider_ours}, without compromising the plasticity loss (FWT) as observed by BiC, Table \ref{tab:CL_perf}. The different ILT versions on the other hand achieve slightly better performances than EWC.

\begin{table}[htbp!]
\begin{adjustbox}{max width=\textwidth}
\begin{tabular}{lcc}\toprule
Method & BWT $\uparrow$ [\%] & FWT $\uparrow$ [\%] \\ \hline \hline
\textit{Sequential} & $-26.60 \pm 17.88$ & $-25.10 \pm 17.44$ \\
\textit{EWC} & $-27.48 \pm 15.00$ & $-29.65 \pm 18.32$ \\
\textit{RWalk} & $-22.79 \pm 18.94$ & $-51.48 \pm 13.23$ \\
\textit{ILT$_{\text{KD}}$} & $-27.86 \pm 11.85$ & $-26.79 \pm 21.32$ \\
\textit{ILT$_{\text{MSE}}$} & $-25.56 \pm 14.90$ & $-34.26 \pm 18.94$ \\
\textit{ILT$_{\text{KD, MSE}}$} & $-16.59 \pm 12.77$ & $-34.58 \pm 17.73$ \\
\textit{BiC} & $\mathbf{-0.56 \pm 7.42}$ & $-29.01 \pm 7.42$ \\
\textit{Rehearsal} & $-6.60 \pm 7.02$ & $-24.23 \pm 15.58$ \\
\hline
\textit{Atlas Replay} & $-8.13 \pm 7.68$ & $\mathbf{-18.26 \pm 15.22}$ \\
\bottomrule
\end{tabular}
\end{adjustbox}
\caption{CL performance indicated by BWT and FWT along with standard deviation.}
\label{tab:CL_perf}
\end{table}

With the default U-Net specifications of VoxelMorph, the average mean segmentation Dice is about $61\%$. However, a significant improvement in our proposed method can be clearly observed. 

\subsection{Qualitative temporal evaluation}
\label{ssec:qual}
To analyze the robustness of our proposed method qualitatively, we visualize segmentation masks in Figure \ref{fig:trans_res}.

\begin{figure*}[htbp!]
    \centering
    \includegraphics[trim=0 6.5cm 2cm 0,clip,width=0.9\textwidth]{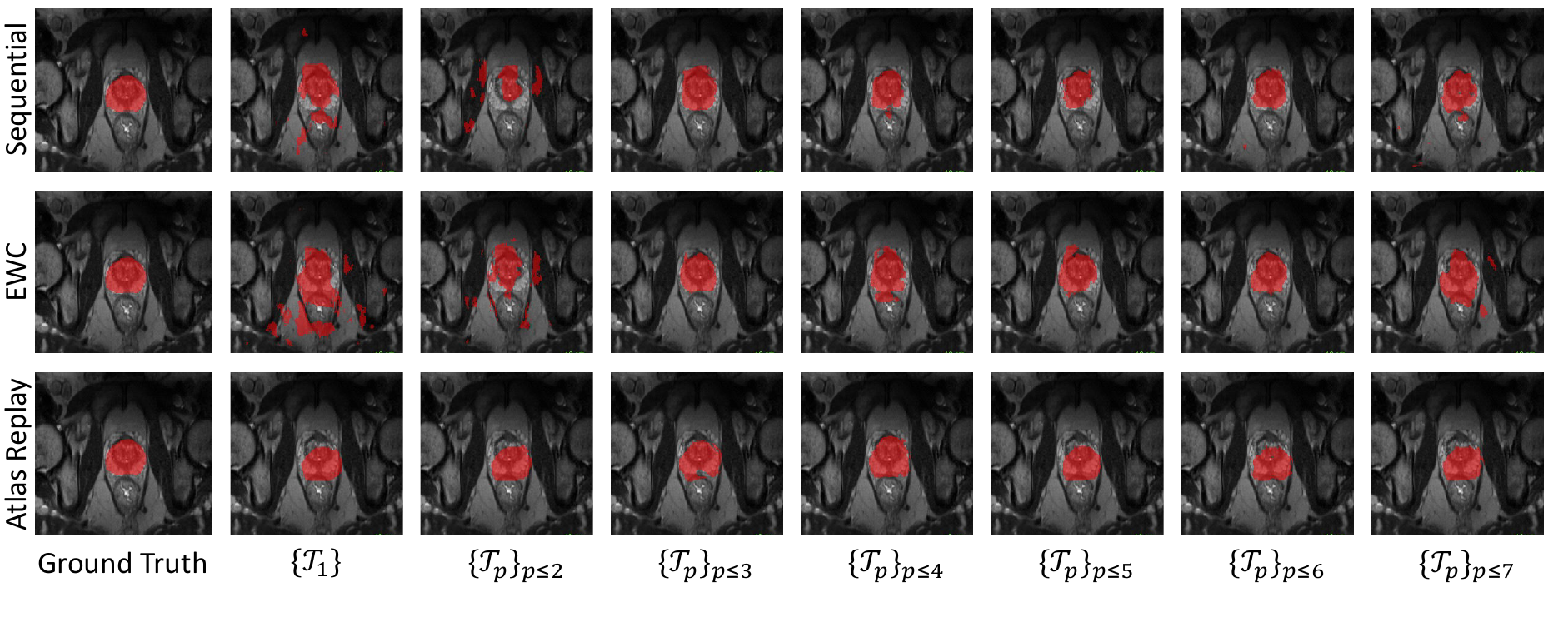}
    \caption{Temporal analysis for sequential, EWC, and Atlas Replay using Case 3, Slice 19 from $\mathcal{T}_3$.}
    \label{fig:trans_res}
\end{figure*}

Atlas Replay consistently produces coherent segmentation masks, irrespective of the training stage. Sequential training and EWC on the other hand produce low-quality segmentations until the network is trained on the particular stage 3 $\left(\{\mathcal{T}_{p}\}_{p \leq 3}\right)$. In particular, the low performance on later tasks shows the effect of catastrophic forgetting, where the network adapts too strongly to the later training data. Atlas Replay is neither too rigid nor plastic, as is outputs robust predictions for data from both early and later training stages.

The image shows the impact of selecting the correct prototype during inference for the effectiveness of the registration method. Using a prototype that was built from a non-coil dataset ($\mathcal{P}_{A}$) to perform registration with a coil-acquired sample ($\mathcal{T}_{2}$) is expected to have lower performance as the geometric shape of the prostate changes significantly depending on the type of coil. Since the coil type is a central aspect of the examination that can be easily recognized, it is to be expected that the user applying the algorithm knows the type of coil that was used during acquisition and can select the correct prototype accordingly. If this were not known, a simple solution would be to register the desired sample across all prototypes and then use the prototype leading to the best registration performance in terms of NCC or MSE.

\subsection{Effectiveness of privacy preserving prototypes}
\label{ssec:user}
In order to evaluate the effectiveness of our privacy-preserving prototypes, we conducted a user study involving two senior radiologists with more than 10 years of experience (RAD) and two computer scientists (CS). The study aimed to assess the participants' ability to correctly select the used subject from a set of three patient scans, with only one scan being the correct match for the shown prototype. This selection process was performed five times for each prototype.

To measure the performance of the participants, we report four key metrics: sensitivity, specificity, precision (positive predicted value), and Matthews Correlation Coefficient (MCC). Sensitivity refers to the ability to correctly identify the true positive cases, i.e., correctly selecting the matching patient scan. Specificity measures the ability to correctly identify the true negative cases, meaning to correctly exclude the non-matching patient scans. Precision represents the proportion of correctly selected matching patient scans out of the total selected matching scans. MCC on the other hand provides an overall assessment of the classification performance, taking into account both true positive and true negative rates. Table \ref{tab:user} shows the average results of our user study for the radiologists and computer scientists including the standard deviation and random chance.

\begin{table}[htp]
\centering
\begin{adjustbox}{max width=\linewidth}
{\begin{tabular}{l|cccc}
\toprule 
Group & Sensitivity $\uparrow$ [\%] & Specificity $\uparrow$ [\%] & Precision $\uparrow$ [\%] & MCC $\uparrow$ [\%] \\ \midrule \midrule

random & \multirow{2}{*}{--} & \multirow{2}{*}{--} & \multirow{2}{*}{$33.33 \left(\frac{1}{3}\right)$} & \multirow{2}{*}{--}\\
chance &  & & &\\
\hline
RAD & $\mathbf{47.50 \pm 7.50}$ & $\mathbf{74.00 \pm 4.00}$ & $\mathbf{47.50 \pm 7.50}$ & $\mathbf{21.50 \pm 14.50}$\\
CS & $45.00 \pm 10.00$ & $73.00 \pm 5.00$ & $45.00 \pm 10.00$ & $18.00 \pm 15.00$\\
\bottomrule
\end{tabular}}
\end{adjustbox}
\caption{Results of our user study evaluation on privacy-preserving prototypes compared to random chance. The evaluation included participation from both radiologists and computer scientists.}
\label{tab:user}
\end{table}

Table \ref{tab:user} clearly shows that correctly selecting the used subject from a set of three samples, along with the prototype, proved to be very challenging for senior radiologists as well as computer scientists. The results illustrate that possessing technical knowledge about the process of prototype building has minimal impact on selecting the correct samples as the classification rates are very similar to the ones from the radiologists. Such a difficulty in correctly identifying the matching patient scan demonstrates the effectiveness of our privacy-preserving prototype building approach. The average precision achieved by the radiologists was $47.50\%$, indicating the complexity of the task and the privacy preservation capabilities of our prototypes.

\subsection{Inter-prototype performance}
\label{ssec:ablation_prot}

\begin{figure}[htbp!]
    \centering
    \includegraphics[width=\textwidth]{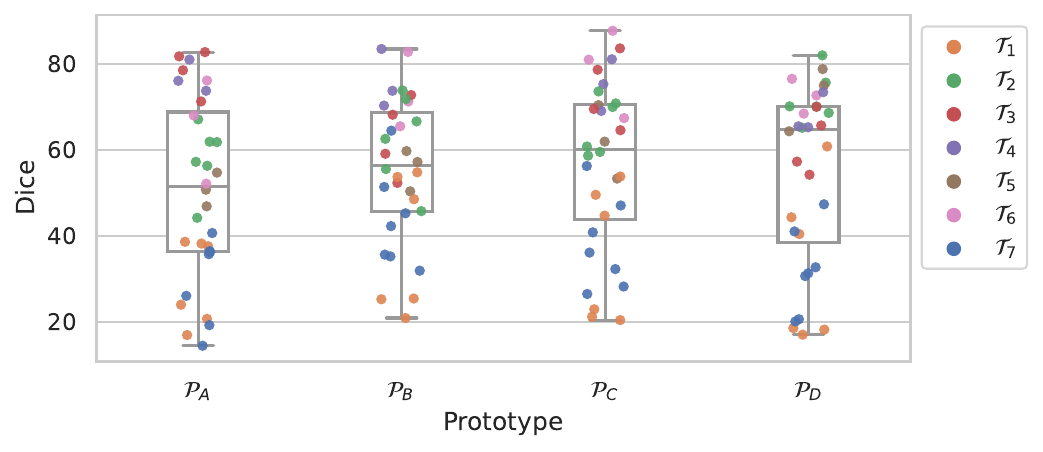}
    \caption{Combination of scatter and box plot showing performance distribution and change in Dice for every task based on the used prototype during inference with the final model trained sequentially on all datasets.}
    \label{fig:prot_abl}
\end{figure}

To assess the inter-prototype performance, we trained sequentially on all seven tasks using Atlas Replay and utilized the final model to analyze the influence of prototypes during inference (Figure \ref{fig:prot_abl}). Specifically, for each sample from the test sets we performed registration using all four prototypes.


Figure \ref{fig:prot_abl} illustrates the distribution of Dice for every task based on the utilized prototype during inference with the sequentially trained model. This visualization effectively shows how the choice of prototype during the registration process influences the performance observed in inference across different tasks.

\subsection{Loss ablation study}
\label{ssec:loss}
To determine the optimal setup for atlas-based segmentation, we conduct an ablation study where we modify the traditional registration loss proposed in the VoxelMorph paper \cite{balakrishnan2019voxelmorph}. This study allows us to identify the most effective VoxelMorph configuration. The networks are trained on the joint dataset $\bigcup_{p=1}^{\lvert \Omega_{\mathcal{T}} \rvert} \mathcal{T}_{p}$ and the evaluation is performed across all seven datasets $\{\mathcal{T}_{p}\}_{p \leq \lvert \Omega_{\mathcal{T}} \rvert}$.

\begin{table*}[htp]
\begin{adjustbox}{max width=\linewidth}
{\begin{tabular}{lccccccc}
\toprule
\multirow{2}{*}{Ablation $\left(\mathcal{L}_{reg}, \ref{align:loss}\right)$} & \multicolumn{7}{c}{Dice $\uparrow{ } \pm{ } $ $\sigma \downarrow $ {[}\%{]}} \\ \cmidrule{2-8}
& $\mathcal{T}_1$ & $\mathcal{T}_2$ & $\mathcal{T}_3$ & $\mathcal{T}_4$ & $\mathcal{T}_5$ & $\mathcal{T}_6$ & $\mathcal{T}_7$ \\ \midrule \midrule
$\mathcal{L}_{NCC} + 0 \cdot \mathcal{L}_{CE} + \mathcal{L}_{smooth}$ & $37.03 \pm 17.05$ & $62.63 \pm 1.97$ & $66.56 \pm 6.06$ & $69.78 \pm 2.08$ & $58.48 \pm 6.94$ & $63.73 \pm 8.74$ & $38.84 \pm 11.86$ \\
$\mathcal{L}_{NCC} + 1 \cdot \mathcal{L}_{CE} + \mathcal{L}_{smooth}$ & $56.01 \pm 15.05$ & $68.95 \pm 3.46$ & $70.83 \pm 3.36$ & $73.80 \pm 1.27$ & $\mathbf{60.18 \pm 8.94}$ & $63.28 \pm 11.71$ & $57.89 \pm 10.83$ \\
$\mathcal{L}_{NCC} + 2 \cdot \mathcal{L}_{CE} + \mathcal{L}_{smooth}$ & $\mathbf{70.55 \pm 6.32}$ & $\mathbf{73.29 \pm 3.65}$ & $\mathbf{71.39 \pm 1.94}$ & $\mathbf{79.84 \pm 3.49}$ & $60.07 \pm 4.88$ & $\mathbf{65.36 \pm 8.89}$ & $\mathbf{69.47 \pm 6.70}$ \\ \hline
$\mathcal{L}_{CE}$ (U-Net joint)  & $79.62 \pm 2.50$ & $77.01 \pm 11.10$ & $78.95 \pm 6.61$ & $79.72 \pm 4.13$ & $83.49 \pm 1.97$ & $82.45 \pm 1.98$ & $80.34 \pm 1.94$ \\
\bottomrule
\end{tabular}}
\end{adjustbox}
\caption{VoxelMorph ablations trained on joint prostate data with different loss variations (top) and U-Net joint results (bottom).}
\label{tab:ablation}
\end{table*}

Table \ref{tab:ablation} shows that weighting the Cross-Entropy loss twice leads to the best segmentation performance across all tasks. A network trained with no segmentation loss term, i.e. only for registration, achieves a limited contribution in terms of segmentation performance. Given this insight, we use the double weighted Cross-Entropy loss setup for all our VoxelMorph related experiments as shown in Equation \ref{align:loss}.

\subsection{End-to-end segmentation vs. atlas-based segmentation}
\label{ssec:gener}
To better assess the generalizability of the models, we train U-Net and Atlas Replay networks for every dataset $\mathcal{T}_{p} \subset \Omega_{\mathcal{T}}$ and validated them across all datasets.

\begin{figure}[htb!]
    \centering
    \includegraphics[width=\textwidth]{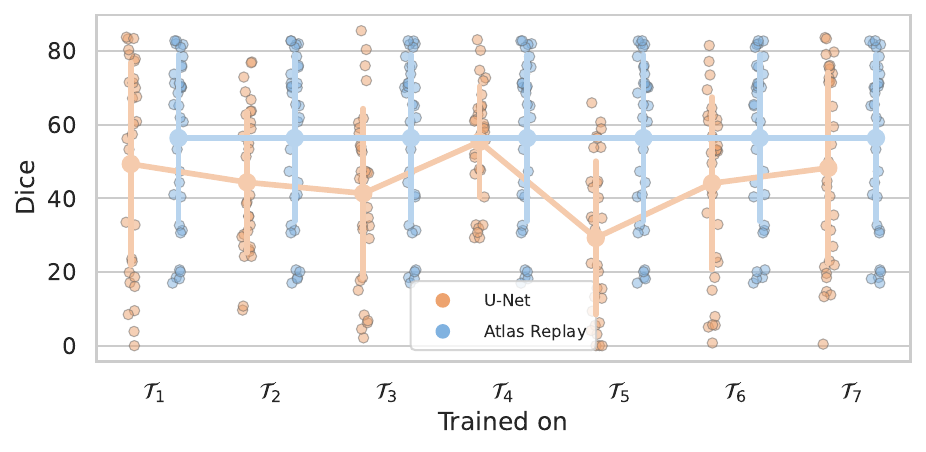}
    \caption{Scatter plot showing the performance distribution in Dice when trained on one dataset and evaluated across all seven datasets from the prostate corpus $\Omega_{\mathcal{T}}$; trend line is based on mean Dice performance across all evaluated samples.}
    \label{fig:scatter}
\end{figure}

Figure \ref{fig:scatter} illustrates the difference between an end-to-end segmentation approach (U-Net) and Atlas Replay in terms of generalizability and segmentation performance. Using the proposed registration-based approach, we obtain robust results regardless of what dataset is used for training. The U-Net performance on the other hand is clearly dependent on the training dataset, as indicated by the strong distribution shifts in the figure, i.e. lack of generalizability. $\mathcal{T}_5$, for instance, shows a significant performance deterioration. Besides an increase in generalizability, the performance of the registration-based method is also higher, as shown by the mean Dice trend line, which is consistently above that of the U-Net. These results show the increased versatility and robustness of Atlas Replay even before continual learning is performed.

\subsection{Hyperparameter search}
\label{ssec:param}
For every used end-to-end CL method, a hyperparameter search was performed using four different settings. The setup with the highest mean performance in terms of Dice, BWT, and FWT together was selected for our experiments. Table \ref{tab:ps} shows the results for each hyperparameter setting. The highest values for a method and the best parameter settings are marked in bold.

\begin{table}[htbp!]
\centering
\begin{adjustbox}{max width=\linewidth}
{\begin{tabular}{lccccc}
\toprule
\multirow{2}{*}{Method} & \multicolumn{5}{c}{Prostate} \\ \cmidrule{2-6} & Fixed params & Tuned param & Dice $\uparrow$ [\%] & BWT $\uparrow$ [\%] & FWT $\uparrow$ [\%] \\ \midrule \midrule
\multirow{4}{*}{$\text{EWC}$} & \multirow{4}{*}{--} & $\lambda = 0.4$ & $42.44 \pm 20.67$ & $-28.47 \pm 12.35$ & $-28.87 \pm 19.48$  \\
& & $\lambda = 1.1$ & $44.83 \pm 20.86$ & $-29.28 \pm 14.82$ & $\mathbf{-27.47 \pm 16.29}$ \\
& & $\lambda = 1.7$ & $44.03 \pm 20.49$ & $-30.96 \pm 15.33$ & $-30.59 \pm 16.66$ \\
& & $\mathbf{\lambda = 2.2}$ & $\mathbf{46.06 \pm 18.46}$ & $\mathbf{-27.48 \pm 15.00}$ & $-29.65 \pm 18.32$ \\
\midrule
\multirow{4}{*}{$\text{RWalk}$} & \multirow{4}{*}{\shortstack{$\alpha = 0.9,$ \\ update $= 20$}} & $\lambda = 0.4$ & $29.94 \pm 23.59$ & $-29.01 \pm 12.84$ & $\mathbf{-46.69 \pm 13.29}$ \\
& & $\lambda = 1.1$ & $27.80 \pm 20.07$ & $-25.91 \pm 17.21$ & $-50.06 \pm 6.757$ \\
& & $\mathbf{\lambda = 1.7}$ & $\mathbf{30.58 \pm 21.38}$ & $-22.79 \pm 18.94$ & $-51.48 \pm 13.23$ \\
& & $\lambda = 2.2$ & $30.48 \pm 23.26$ & $\mathbf{-21.43 \pm 10.50}$ & $-54.31 \pm 11.07$ \\
\bottomrule \\
\end{tabular}}
\end{adjustbox}
\caption{Results of the hyperparameter search considered for the two used end-to-end CL methods; mean Dice, BWT and FWT over all tasks including standard deviation [in \%]; highest values and best parameter setup are marked in bold.}
\label{tab:ps}
\end{table}

The results clearly show the addressed trade-off problem from Figure \ref{fig:trade}. Depending on how the hyperparameter(s) are set, the trade-off between plasticity and knowledge preservation varies. The network either performs well in terms of maintaining knowledge over time -- increased $\lambda$ for EWC -- or the results are very similar no matter how the hyperparameter is set -- $\lambda$ for RWalk. 

\section{Conclusion}
We introduce \textit{Atlas Replay}, a robust atlas-based segmentation technique for continuous training in clinical environments with data drift. We evaluate our approach on seven prostate segmentation scenarios and show that it outperforms state-of-the-art continual learning methods by maintaining knowledge from early stages without compromising model plasticity. \emph{Atlas Replay} is more generalizable than U-Net models even when trained statically with only data from one site, with a performance difference of $15\%$ on average. We additionally introduced a prototype-building method with initial privacy preservation from a human perspective that can be leveraged to maintain structural information over time resulting in a proper balance between rigidity and plasticity for CL setups. Future work should address more stringent privacy preservation for prototypes and a self-growing atlas technique in combination with proper prototype adjustments over time to further push the potential of registration for segmentation in terms of CL. By releasing our code base along with instructions and pre-trained networks, we hope to amplify and inspire CL research beyond end-to-end models that incorporates atlas-based segmentation for medical settings.

\section{Acknowledgements}
This work was (partially) supported by RACOON (NUM), under BMBF grant number 01KX2021 and EVA-KI, under BMG grant number ZMVI1- 2520DAT03A.

\onecolumn
{\centering{\Large \bf Supplementary Material \par}}
\setcounter{section}{0}
\vspace{0.25cm}
\section{End-to-end vs. atlas-based segmentation}
\label{supp}
\begin{minipage}{7cm}
Table \ref{tab:baselines} provides the Dice scores with standard deviation for every trained baseline evaluated across all tasks which were used to create Figure \ref{fig:scatter} from the main manuscript.
\end{minipage}%

\begin{table*}[h!]
\begin{center}
\begin{adjustbox}{max width=\linewidth}
{\begin{tabular}{ccccccccc}
\toprule
& \multirow{2}{*}{Baselines} & \multicolumn{7}{c}{Dice $\uparrow{ } \pm{ } $ $\sigma \downarrow $ {[}\%{]}} \\ \cmidrule{3-9}
& & $\mathcal{T}_1$ & $\mathcal{T}_2$ & $\mathcal{T}_3$ & $\mathcal{T}_4$ & $\mathcal{T}_5$ & $\mathcal{T}_6$ & $\mathcal{T}_7$ \\ \midrule \midrule
\parbox[t]{2mm}{\multirow{7}{*}{\rotatebox[origin=c]{90}{U-Net}}} & $\mathcal{T}_{1}$ & $\mathbf{75.38 \pm 5.96}$ & $37.82 \pm 15.47$ & $17.04 \pm 4.83$ & $42.02 \pm 23.75$ & $6.66 \pm 6.86$ & $50.86 \pm 1.78$ & $76.14 \pm 5.66$ \\
& $\mathcal{T}_{2}$ & $36.36 \pm 19.55$ & $\mathbf{70.25 \pm 6.30}$ & $25.81 \pm 9.26$ & $52.21 \pm 15.53$ & $50.30 \pm 8.01$ & $36.78 \pm 7.98$ & $37.01 \pm 10.70$ \\
& $\mathcal{T}_{3}$ & $21.12 \pm 17.64$ & $53.55 \pm 6.23$ & $\mathbf{78.50 \pm 5.04}$ & $55.44 \pm 2.65$ & $37.54 \pm 6.98$ & $50.12 \pm 5.39$ & $18.94 \pm 9.90$ \\
& $\mathcal{T}_{4}$ & $46.48 \pm 17.13$ & $65.69 \pm 6.29$ & $60.23 \pm 2.08$ & $\mathbf{66.44 \pm 11.89}$ & $49.19 \pm 13.37$ & $65.73 \pm 10.23$ & $44.92 \pm 11.52$ \\
& $\mathcal{T}_{5}$ & $9.61 \pm 7.19$ & $48.44 \pm 8.02$ & $29.82 \pm 5.24$ & $40.26 \pm 18.72$ & $\mathbf{59.55 \pm 4.97}$ & $38.21 \pm 4.85$ & $7.87 \pm 7.35$ \\
& $\mathcal{T}_{6}$ & $22.05 \pm 19.25$ & $54.02 \pm 10.05$ & $58.41 \pm 4.21$ & $59.37 \pm 4.41$ & $52.10 \pm 1.99$ & $\mathbf{77.42 \pm 3.27}$ & $22.13 \pm 17.11$ \\
& $\mathcal{T}_{7}$ & $73.52 \pm 7.57$ & $37.38 \pm 9.17$ & $20.28 \pm 3.27$ & $32.83 \pm 22.86$ & $10.75 \pm 7.59$ & $45.11 \pm 3.84$ & $\mathbf{76.12 \pm 3.83}$ \\ \cmidrule{1-9}
\parbox[t]{2mm}{\multirow{7}{*}{\rotatebox[origin=c]{90}{Atlas Replay}}} & $\mathcal{T}_{1}$ & $\mathbf{70.50 \pm 4.63}$ & $56.02 \pm 4.78$ & $72.87 \pm 6.30$ & $73.79 \pm 1.55$ & $34.81 \pm 22.70$ & $63.69 \pm 9.80$ & $73.02 \pm 3.63$ \\
& $\mathcal{T}_{2}$ & $44.27 \pm 18.47$ & $\mathbf{72.30 \pm 6.05}$ & $73.92 \pm 3.50$ & $75.16 \pm 0.91$ & $53.73 \pm 13.33$ & $66.62 \pm 7.50$ & $43.99 \pm 9.74$ \\
& $\mathcal{T}_{3}$ & $34.30 \pm 7.69$ & $34.69 \pm 6.41$ & $\mathbf{73.11 \pm 4.42}$ & $71.33 \pm 2.28$ & $23.81 \pm 11.07$ & $44.28 \pm 8.05$ & $31.97 \pm 7.77$ \\
& $\mathcal{T}_{4}$ & $23.20 \pm 9.93$ & $23.61 \pm 7.13$ & $63.36 \pm 7.56$ & $\mathbf{78.40 \pm 0.72}$ & $20.57 \pm 7.71$ & $37.11 \pm 4.82$ & $19.55 \pm 5.34$ \\
& $\mathcal{T}_{5}$ & $32.79 \pm 15.13$ & $53.30 \pm 2.42$ & $72.87 \pm 5.67$ & $74.55 \pm 2.59$ & $\mathbf{60.31 \pm 3.02}$ & $58.50 \pm 9.19$ & $30.31 \pm 8.77$ \\
& $\mathcal{T}_{6}$ & $34.66 \pm 15.19$ & $44.05 \pm 6.22$ & $68.91 \pm 3.17$ & $71.48 \pm 4.08$ & $39.16 \pm 15.10$ & $\mathbf{64.97 \pm 13.40}$ & $33.47 \pm 8.32$ \\
& $\mathcal{T}_{7}$ & $69.02 \pm 5.74$ & $53.63 \pm 8.64$ & $74.94 \pm 5.43$ & $73.69 \pm 1.35$ & $36.15 \pm 16.09$ & $59.41 \pm 10.48$ & $\mathbf{67.73 \pm 5.20}$ \\
\bottomrule
\end{tabular}}
\end{adjustbox}
\caption{Results for all baseline networks trained on every task individually and evaluated across all tasks; Bold values indicate the performance of the baseline on the validation set of the task it has been trained on.}
\label{tab:baselines}
\end{center}
\end{table*}

\twocolumn
%
%
%
\bibliographystyle{ieee_fullname}
\bibliography{literature}

\end{document}